\pdfoutput=1

\documentclass[11pt]{article}

\usepackage{acl}

\usepackage{times}
\usepackage{latexsym}
\usepackage{graphicx}
\usepackage{svg}
\usepackage[T1]{fontenc}

\usepackage[utf8]{inputenc}

\usepackage{microtype}
\usepackage{booktabs}

\title{PHEMEPlus: Enriching Social Media Rumour Verification with External Evidence} 

\author{$^1$John Dougrez-Lewis, $^{2,3}$Elena Kochkina, $^{1,3}$Miguel Arana-Catania, $^{1,2,3}$Maria Liakata, $^{1,3}$Yulan He \\
  $^1$Department of Computer Science, University of Warwick, UK\\
  $^2$Queen-Mary University of London, UK\\
  $^3$The Alan Turing Institute, UK\\
  \texttt{\{j.Dougrez-Lewis,miguel.arana-catania,yulan.he\}@warwick.ac.uk}\\ \texttt{\{m.liakata,e.kochkina\}@qmul.ac.uk}
  }

\begin{document}
\maketitle
\begin{abstract}
Work on social media rumour
verification utilises signals from posts, their propagation and users involved. Other lines of work target identifying and fact-checking claims based on information from Wikipedia, or trustworthy news articles without considering social media context.  
However works combining the information from social media with external evidence from the wider web are lacking. 
To facilitate research in this direction, we release a novel dataset, PHEMEPlus\footnote{https://github.com/JohnNLP/PhemePlus}, an extension of the PHEME benchmark, which contains social media conversations as well as relevant external evidence for each rumour. 
We demonstrate the effectiveness of incorporating such evidence in improving rumour verification models.
Additionally, as part of the evidence collection, we evaluate various ways of query formulation to identify the most effective method.
 


\end{abstract}

\section{Introduction}


The harm and prevalence of online misinformation made research into automated methods of information verification an important and active research area. This includes various tasks like fact-checking,  social media rumour detection, stance classification and verification. 
In this work we are concerned with social media rumour verification, the task of identifying whether a rumour (i.e check-worthy claim circulating on social media whose veracity status is yet to be verified~\cite{zubiaga}), is \emph{True}, \emph{False} or \emph{Unverified}. 

Although a significant amount of work has been done towards evaluating the veracity of social media rumours~\cite{zubiaga2016analysing,tw1516,identify_them,SAVED}, there is still a dearth of works and datasets combining the information from social media with external evidence from the wider web. While recent works focusing on rumours around the COVID-19 pandemic have been collecting data from a wide range of sources from news and social media to scientific publications ~\citep{cui2020coaid,zhou2020recovery,wang2020cord}, these are not sufficient for the creation of generalisable verification models as they only focus on a single topic. 
At the same time works on fact-checking, which do not focus on social media content, but use claims from debunking websites~\citep{015,006}, as well as recent work by \citet{li2021meet} have shown the benefits of utilising stance of evidence for verification. 

Here we aim to further enable research in this direction and release an enriched version of a popular benchmark dataset PHEME~\citep{zubiaga2016analysing} with timely evidence for each of the rumours, obtained from a wide range of web sources. 

Although a few works use web search for evidence retrieval~\cite{011,015}, to our knowledge, only the work of \citet{021} 
touches upon the topic of the search query formulation.
Here we analyse several query formulation strategies to find the most effective one. 



In this work we make the following contributions: 
\begin{itemize}




\item We collect and release the PHEMEPlus dataset of Twitter rumour conversations with the relevant heterogeneous evidence retrieved from the web to facilitate research on combining multiple sources of information for social media rumour verification.

\item 
We investigate approaches towards search query formulation for evidence retrieval, together with evaluation metrics for the quality of evidence retrieved.

\item We demonstrate the effectiveness of incorporating external evidence into rumour veracity classification models.

\end{itemize}

\section{Related work}

\subsection{Existing Veracity Classification Datasets}
Among existing datasets for veracity classification we can broadly discern two categories: (1) focusing on claims arising from social media in the form of posts~\citep{zubiaga2016analysing,tw1516} and (2) focusing on manually formulated claims, either created specifically for a task~\citep{fever}, or  consisting of titles from news or debunking websites~\citep{wang2017liar,alhindi2018your,015,006}.
These different types of claims present different challenges for verification models and evidence retrieval systems. In particular social media posts often use non-standard grammar, hashtags and have typos (intentional or otherwise). It can be crucial to process claims directly from social media to enable early-stage misinformation detection as rumours often start spreading on social media, later making it into the mainstream media. Only a few datasets incorporate both social media and evidence from the web, however these often focus on a very limited number of sources of evidence or a single topic~\citep{dai2020ginger,cui2020coaid}. 
One of such datasets is FakeNewsNet~\cite{shu2018fakenewsnet} incorporating \emph{fake} and \emph{true} news articles from fact-checking websites PolitiFact\footnote{https://www.politifact.com/} and GossipCop\footnote{https://www.suggest.com/}. Articles are further augmented with users' posts on Twitter pertaining to them but not including full conversation structure.  FakeHealth~\cite{dai2020ginger} is a similarly constructed dataset based on health-related news articles labelled by the Health News Review\footnote{https://www.healthnewsreview.org/}, including Twitter users' replies and profiles. 
\citet{barron2020checkthat} organised shared tasks for automatic identification and verification of claims in social media. Apart from tasks on check-worthiness estimation for tweets and verified claim retrieval, they also released tasks for supporting evidence retrieval and claim verification. However, the tasks mainly focused on misinformation about COVID-19 and the latter tasks were only offered in Arabic. 

In light of the wave of misinformation associated with COVID-19 pandemic researchers have been collecting relevant datasets of scientific publications, news articles and their headlines, social media posts and claims about COVID-19~\citep{shaar2020overview,dharawat2020drink,zhou2020recovery,li2020mm,memon2020characterizing,hossain2020covidlies,barron2020checkthat}. One of the most relevant work to ours is COAID \cite{cui2020coaid}, a large-scale dataset containing COVID-19 related news articles as well as social media posts. While these are rich resources, which enable further research against misinformation, they are insufficient for training generalisable models as they solely focus on one topic. 




In this work we have augmented the PHEME dataset, a popular benchmark dataset for social media rumour verification, it contains rumours expressed via Twitter posts with full conversation threads from several news-breaking events on different topics. This dataset is set up to imitate realistic scenarios as (1) it was collected as the events were unfolding and then rumour stories were identified and annotated by a professional journalist as opposed to collecting tweets based on existing fact-checks as in \citet{tw1516}; and (2) the evaluation is performed in on events unseen during training. We augment it with evidence articles from across the web to give it access to an unlimited set of resources. To preserve the realistic scenario of verifying emerging rumours, all of our evidence is restricted to articles indexed by Google no later than the day on which the rumour was posted to Twitter.

\subsection{Social Media Rumour Verification Models Using External Information}
Social media rumour verification 
models use various types of information available on social media platform: text of rumourous posts and responses~\citep{SAVED}, user information and connections~\citep{khoo2020interpretable}, propagation patterns~\citep{ma2018rumor}. However, still only few works incorporate external evidence.

 \citet{021} proposed the iFACT framework that extracts claims from tweets pertaining to major events. For each claim, it collects evidence from web search and estimates the likelihood of a claim being credible. To formulate the search query iFACT uses ClausIE \citep{clausie} to extract (\emph{subject, predicate, object}) triples from tweets. 
 To determine the credibility of the claim iFACT uses features extracted from search results and dependencies between claims.
Here we also experiment with using ClausIE to formulate the search query. 


\begin{table*}[htb]
\centering
\begin{tabular}{lrrrrr}
\toprule
Events & Threads & True & False & Unverified & Relevant Articles \\ \midrule
Charlie Hebdo                    & 458                              & 193                            & 116                          & 149                                  &            3941         \\ 
Sydney Siege                     & 522                             & 382                         & 86                              & 54                                   &             4436        \\ 
Ferguson                         & 284                             & 10                             & 8                           & 266                                 &              2473       \\ 
Ottawa Shooting                  & 470                              & 329                        & 72                              & 69                                   &          4020           \\ 
Germanwings Crash                & 238                            & 94                            & 111                            & 33                                   &          2057           \\ \midrule
Total Threads  &           1972                  &        1008                  &        393                    &       571                       &       16927              \\
\bottomrule
\end{tabular}
\caption{Statistics of the PHEMEPlus dataset by extending the PHEME-5 dataset with retrieved relevant articles. All but 2 rumours have at least 1 associated article.} 
\label{tab:statspheme5}
\end{table*}

\begin{figure*}[ht]
\centering
\includegraphics[width=0.9\linewidth]{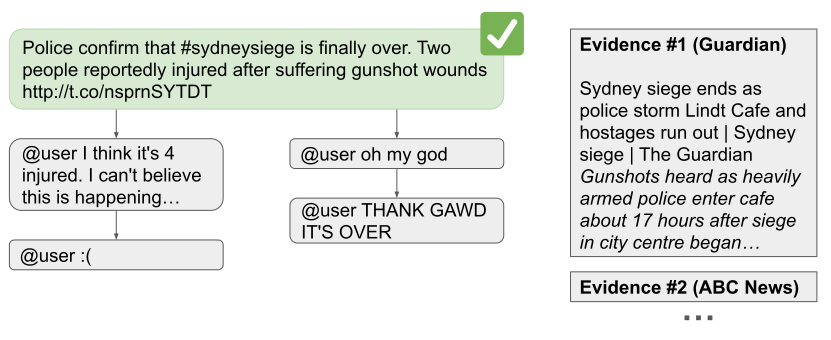}
\caption{\label{data} The PHEMEPlus dataset consists of labelled Twitter rumours, their conversation thread, and corresponding evidence retrieved from the web. 
This is an adapted example.}
\end{figure*}
\citet{li2021meet} propose to improve rumour detection on PHEME dataset by using evidence from Wikipedia. They first train the evidence extraction module on the FEVER dataset and then use it as part of a rumour detection system to get relevant sentences from a Wikipedia dump along with Twitter conversation around a rumour. While being limited by a single source of information, they demonstrate performance improvements  over previous models not using external information. 

In this work we use BERT-based models as strong baselines to demonstrate the effectiveness of incorporating the evidence for social media rumour verification. In future work we will be experimenting with various ways of incorporating it to maximise the benefits.

\section{Augmenting PHEME dataset with External Evidence}

\subsection{Base dataset} \label{loocv}

We chose to extend the PHEME-5 dataset \citep{zubiaga2016analysing}, which consists of Twitter conversations discussing rumours around five real-world events including the Lindt Cafe siege in Sydney and the 2015 Charlie Hebdo terrorist attack.
This dataset is a popular benchmark for rumour verification, it is particularly challenging due to class imbalance and evaluation using leave-one-event-out cross-validation, reflecting a real-world evaluation scenario.
Table \ref{tab:statspheme5} shows the statistics of the PHEMEPlus dataset by extending the original PHEME-5 dataset with retrieved relevant articles. The first four columns show the number of conversation threads in each of the event and each of the classes in the orignal PHEME-5 dataset.  
Figure \ref{data} shows an example entry in the PHEMEPlus dataset, comprised of a rumorous tweet, veracity label, its conversation thread, and relevant evidence retrieved from the web. It is notable that tweets in the conversation thread (and the rumour itself) often contain URLs provided by users which may be useful as a further source of evidence, and that the corresponding evidence is not a part of the original PHEME dataset. \citet{e_thesis} has shown that \emph{True} rumours in PHEME have a higher percentage of URLs attached (55\%) than for \emph{False} (48\%) and \emph{Unverified} (48\%) rumours. For the portion of PHEME with comments annotated for stance, these supplementary URLs were overwhelmingly found in comments \textit{supporting} the source tweet's claim (33\%) as opposed to those, \textit{denying} (8\%), \textit{querying} (6\%), or \textit{commenting} (9\%) on it.


\subsection{Evidence Retrieval through Web search} \label{strategies}

\begin{table*}[htb]
\centering
\begin{tabular}{p{3cm}p{12cm}}
\toprule
\textbf{Original Rumour}:     & \emph{MORE: Massacre suspects believed to have taken hostage and holed up in small industrial town northeast of Paris: <url> \#CharlieHebdo}              \\ \midrule
\textbf{Query Strategy} & \textbf{Query Text} \\ \midrule
Preprocessed  &        before:2015-01-09 MORE : Massacre suspects believed to have taken hostage and holed up in small industrial town northeast of Paris :                  \\
StanfordNLP  &        before:2015-01-09 (Charlie Hebdo) Massacre suspects small industrial town northeast                  \\
ClausIE  &        before:2015-01-09 (Charlie Hebdo) Massacre suspects believed to have taken hostage holed up in small industrial town northeast of Paris                  \\
\bottomrule
\end{tabular}
\caption{Examples of search queries generated by the various search strategies, given the original rumour. In this case, the ClausIE strategy only removes the words "MORE" and "and".}
\label{example_search}
\end{table*}

In order to obtain evidence from the unlimited number of sources we chose to use Web search for evidence retrieval. We choose Google Search as it is one of the most established search engines, and, importantly, allows us to filter results by date. This is crucial as rumours are often resolved and widely debunked in some time following their originating event and the rumourous post, but this information would not be available to the model in a real time evaluation scenario. 

Furthermore, the evidence we retrieve from Google appears robustly reputable, with popular news sources consistently ranking highly in the search results. This is to be expected, since their PageRank system weights heavily websites which are highly cited/referenced by others. Web search results are also more likely to be up-to-date than any corresponding Wikipedia pages regarding a current real world happening, which may not be updated nor appropriately checked for correctness.

For every search we include the term \textsf{(before: date)} at the start of the query to restrict results to articles from before the date the rumourous tweet was posted. For each query we collect the top 5 non-empty results from the web search. 

While Google search 
is able to process various types of queries, from keywords to natural language utterances, we performed a set of experiments to identify the most suitable method of query formulation for our particular task of evidence retrieval for rumours conveyed in Twitter posts. We experiment with queries formulated as (1) natural language sentence, (2) keywords, and (3) \emph{(subject, object, predicate)} triples. For each experiment, we include around 99\% of the PHEME dataset since a few queries did not yield enough non-empty results. 
Although we are aware of some more advanced studies into query expansion and formulation \citep{queries,scells2020automatic}, contributing to these fields is beyond the scope of this paper. Here we aim to demonstrate gains from relatively simple approaches described below towards evidence retrieval.

\subsubsection{Search Strategies}

We experiment with the following search strategies:

\paragraph{Preprocessed}
The search query is the source rumour, obtained from the preprocessed tweet. Our preprocessing entails removing URLs, replacing user mentions with “user” (so as to retain lexical structure), removing hashtags from the end but not the middle (also for lexical structure) and segmenting any compound hashtags. URLs are saved aside since they may have future use as evidence. Hashtags at the end of the tweet (but not others) are also retained, placed in brackets for an "OR" search with the rest of the query. These hashtags in particular are expected to be highly telling of the topic/theme of the tweet, especially when it is otherwise lacking in contextual words.

\paragraph{Shortening with StanfordNLP}
We use Stanza \citep{qi2020stanza} to parse preprocessed tweets. Having obtained a parse tree, words in the following constructs are retained in-place: \{\emph{obl:npmod, compound, advcl, nummod, acl:relcl, nsubj:pass, acl, amod, aux:pass}\}. This combination of constructs was iteratively finetuned until the resultant queries felt similar to the author's own search style, the idea being to replicate the search strategy of an experienced user. Hashtags at the end of tweets are handled as before. 

\paragraph{Shortening with ClausIE}
We use ClausIE \citep{clausie}, a popular subject-relation-object extraction system in the same manner to find (\emph{subject, predicate, object}) triples. These are kept in-place whilst the other words are removed. Hashtags at the end of tweets are retained as before. 


Examples of the search queries formed can be found in Table \ref{example_search}.

\subsubsection{Evaluation metrics}

We devise evaluation metrics to compare the quality of evidence retrieved using different query types, without the need for a rumour verification model in advance. 

\paragraph{URL Words Metric}

URLs frequently contain English words which are representative of the content on their webpage, which we can treat as gold-standard keywords as in ~\citep{ma2016detecting}. To get a goodness score in the range [0,1] we compute the cosine similarity between the words in URLs of retrieved articles and those posted in response to the rumour. Specifically, for each retrieved article, its URL-words are compared with those of each URL in the Twitter comments. The final score is the average of all such cosine similarities across all retrieved articles in the dataset, encoded by Word2Vec \citep{w2v}.

\paragraph{GloVe Metric}

If an article is relevant to a rumour, they will be similar in content. We use GloVe \citep{glove} to calculate the similarity between the first 3 paragraphs of an article and the source rumour, with the title also counting as a paragraph. We use only the first few paragraphs because they seem likely to contain the highest density of relevant information. Cosine similarity scores are calculated between each of these paragraphs and the source rumour, and are averaged to give the article a similarity score. Unknown words with zero vectors are ignored for this purpose, although there is a weakness that some of the most important event-specific words could be unknown.

\paragraph{BERTScore Metric}

This is calculated similarly to the GloVe metric, except that BERTScore \citep{bertscore} is used in its place.



\subsubsection{Evaluating Retrieval Results}

Table \ref{stats2} displays the performance of our search strategies when evaluated via the URL Words, GloVe, and BERT evaluation metrics. These results suggest that searching for the preprocessed tweet may be the best way to get relevant background information from the web, as opposed to extracting keywords from the tweet. This narrowly surpasses the performance of our ClausIE-based search strategy, which outperforms the StanfordNLP approach. The ClausIE strategy may retain a higher proportion of key grammatical constructs than the latter, which play an unexpectedly important role in Google's search algorithm. This is contrary to the authors' searching intuition, perhaps due to their recent integration of models such as BERT \cite{bert}. 

\begin{table}[hbt!]
\centering
\resizebox{\columnwidth}{!}{%
\begin{tabular}{lccc}
\hline \textbf{Metric} & 
\textbf{Preprocessed} & \textbf{StanfordNLP} & \textbf{ClausIE}  \\ \hline
URL Words & 
\textbf{0.802} & 0.777 & 0.795  \\
GloVe  & 
\textbf{0.661} & 0.651 & 0.660  \\
BERTScore & 
\textbf{0.826} & 0.825 & 0.825  \\
\hline
\end{tabular}}
\caption{\label{stats2} Performance of the search strategies, evaluated by our evaluation metrics. 
}
\end{table}


Although some of the values in Table \ref{stats2} appear close together, it is notable that the results of the different query formulations land in the same order irrespective of the scoring metric used. Furthermore, the score differences between different query formulations become more substantial when taking into account their weak upper and lower bounds derived from using artificially generated `target article' and `random' queries (data not shown).




\subsection{PHEMEPlus dataset}

An example entry of the PHEMEPlus dataset can be found in Figure \ref{data}. The number of articles we retrieved using the \textit{Preprocessed} method can be found in Table \ref{tab:statspheme5}. All but two of the rumours have at least one associated evidence article, up to a maximum of 10. 



\begin{table*}[hbt!]
\centering
\begin{tabular}{lcccc}
\hline \textbf{ } & \textbf{Overall pages} & \textbf{Unique pages}\\ \hline
From web search & 13255 (12008 not-empty) & 3817 (3425 not-empty) \\
From rumour responses & 2160 (1658 not-empty) & 601 (457 not-empty) \\
Overlap & 100 & 102 \\
\hline
\end{tabular}
\caption{\label{stats3}Overlap of retrieved articles with articles from rumour responses.}
\end{table*}

We explore the overlap between the evidence in our resultant PHEMEPlus dataset and the URLs in the Twitter comments responding to the rumours. Table \ref{stats3} shows the overlap between the articles retrieved from web search (using the Preprocessed Only strategy) and those from the Twitter comments. We observe little overlap between articles retrieved from web search and articles retrieved from comments responding to rumours. The latter may thus be a substantially different, potentially less useful, source of evidence due to a high density of social media pages and the likelihood that some of the comments may not be directly responding to the source rumour.


A relatively large proportion of the articles retrieved from responses are deemed "empty", meaning they either have no body-text and/or no title. From this, and manual inspection, we infer that response-URLs are more likely to be social media posts or videos which are prone to missing titles or first paragraphs. 

The overall:unique ratio being similar for both thread and web suggests that the Google results are indeed sensitive to the content of each thread, as opposed to repeatedly giving the same results for a given rumourous event. There is not much overlap between the search results and the Twitter thread, and a large proportion of existing overlap might be explainable by news websites tweeting their news URLs. This is not attributable to overly stringent overlap criteria as the discrepancy between the overall number of articles and the number of articles without duplicates acts as a positive control to this end.

Similar links nearly always result from the same thread, possibly due to the aforementioned news companies. Investigating further, the vast majority (if not all) of the overlap was news articles. Speculatively, it is plausible that most of this overlap came from news websites tweeting their stories, as there are some examples of this in the dataset.

\section{Evaluating the Effectiveness of Evidence for Rumour Verification}

We conduct experiments to evaluate the effectiveness of our retrieved evidence for Twitter rumour veracity classification. 

\subsection{Evidence Sentence Retrieval}
In our PHEMEPlus dataset, each source tweet is paired with up to 10 most relevant retrieved articles. We follow the typical pipeline fact checking approach to further select the 5 most relevant sentences from the articles associated with each source tweet. In order to do this, we use a simple novel approach based on ClausIE \citep{clausie}. The idea is to be able to reliably find relevant sentences whilst not being clobbered by the inevitably rare rumour-specific vocabulary which may not be recognised by many approaches. First, we use ClausIE to extract all relevant subject-predicate-object triples from the retrieved information. We assume these to be the words with the most potential for true relevance to the tweet. Any stop-words contained within are filtered out. For each sentence, a score is assigned based on how many of these important words are also contained in the tweet, penalising both overly long (>20 token) and short (<5 token) sentences as are likely to be either uninformative or unconcise and work poorly with the BERT models. In particular, short sentences are ignored, whereas long sentences lose 2\% of their score for each additional word. Only rumours with enough evidence to extract 5 sentences as above are used (99\% of them) in our experiments. The top 5 such sentences are paired with each source tweet and are fed into a rumour classification model for veracity assessment.



\begin{table*}[h]
\centering
\begin{tabular}{lcccccccccccc}
\hline \textbf{BERT} & \vline  & \textbf{Ch} & \textbf{Fe} & \textbf{Ge} & \textbf{Ot} & \textbf{Sy} &\vline & \textbf{False}  & \textbf{True}  & \textbf{Unv} &\vline  & \textbf{MacroF1}\\ \hline
Rumour + Ev. &\vline  & \textbf{0.317} & \textbf{0.174} & 0.213 & \textbf{0.406} & 0.318 &\vline & \textbf{0.221} & 0.549 & \textbf{0.265} &\vline  & \textbf{0.345} \\
Rumour &\vline  & 0.306 & 0.134 & \textbf{0.315} & 0.345 & \textbf{0.320} &\vline & 0.209 & 0.562 & 0.242 &\vline & 0.338 \\
Evidence &\vline & 0.268 & 0.045 & 0.264 & 0.370 & 0.307 &\vline & 0.140 & \textbf{0.645} & 0.099 &\vline & 0.295\\
\hline \textbf{RoBERTa} & \vline  & & & &  & &\vline &   &  &&\vline  & \\ \hline
Rumour + Ev. &\vline  & \textbf{0.306} & \textbf{0.183} & \textbf{0.383} & 0.368 & \textbf{0.347} &\vline & \textbf{0.384} & 0.600 & \textbf{0.279} &\vline  & \textbf{0.421} \\
Rumour &\vline  & 0.290 & 0.113 & 0.260 & \textbf{0.420} & 0.309 &\vline & 0.211 & 0.549 & 0.232 &\vline & 0.331 \\
Evidence &\vline & 0.288 & 0.028 & 0.252 & 0.335 & 0.327 &\vline & 0.145 & \textbf{0.611} & 0.144 &\vline & 0.301\\
\hline \textbf{NLI-SAN} & \vline  &  & &  &  & &\vline &  &  &  &\vline  & \\ \hline
Rumour + Ev. &\vline  & \textbf{0.354} & \textbf{0.256} & \textbf{0.365} & \textbf{0.591} & \textbf{0.458} &\vline & \textbf{0.186} & \textbf{0.480} & \textbf{0.250} &\vline  & \textbf{0.405} \\
\hline
\end{tabular}
\caption{\label{best_results} Per-event and per-fold F1 scores from the BERT, RoBERTa, and NLI-SAN models. The 2-letter column headings abbreviate the names of individual rumourous events in PHEME (as in Table \ref{tab:statspheme5}).} 
\end{table*}

\subsection{Veracity Classification Models}

We compare the performance of several veracity classification models in three input scenarios: (1) rumour (i.e., source tweet) alone, (2) evidence (i.e., extracted sentences) alone and (3) rumour concatenated with the evidence (extracted sentences). The classification models chosen include pre-trained language models such as BERT \citep{bert} and RoBERTa \citep{roberta}, and a model making use of natural language inference results between a source rumour and its related evidence sentence.



\paragraph{BERT-based approaches} 

We train BERT-based models including BERT and RoBERTa followed by a single softmax layer for rumour verification. Each pair of a rumour and a piece of relevant evidence sentence is concatenated as input to the model. The final predictions were determined by majority voting. These particular models are chosen because flavours of BERT have previously achieved state-of-the-art results in many natural language processing tasks. 



\paragraph{Self-Attention Network based on Natural Language Inference (NLI-SAN)}

This method uses not only the representation of rumour and evidence like the previous methods, but also the Natural Language Inference (NLI) relationship between them.

First each rumour is paired with each of the evidence sentences and is fed into the RoBERTa-large-MNLI\textsuperscript{\ref{huggface}} model to generate the NLI relation triplet representing the \emph{contradiction}, \emph{neutrality}, and \emph{entailment} probabilities. The rumour-sentence pair is also fed into the RoBERTa-large\footnote{\label{huggface}\url{https://huggingface.co/}} model to generate the contextual representation. Both outputs are then combined using a self-attention network in which the NLI relation triplet is used as the query, while the contextual representation is used as the key and value. Afterwards, all the outputs are concatenated into a single output that is passed through a Multi-Layer Perceptron (MLP) and a Softmax layer that generates the final veracity classification value.






Since this approach relies on the inference relationship between rumour and evidence, we will only compare it with the other models if both elements are available, and thus only one result is shown in Table \ref{best_results}.



\subsection{Experimental Setup}

Experiments were performed using 5-fold leave-one-out-cross-validation with each of PHEME's rumourous events being a fold, as is customary for this dataset (see Section \ref{loocv}).  
We will release the code used to collect the evidence and to perform experiments on GitHub.

For the training of the aforementioned models, the inputs are padded and truncated to the longest sequence. Cross-entropy is used as the loss function. The optimizer used is AdamW \citep{loshchilov2017decoupled} with $\beta_1=0.9$, $\beta_2=0.999$, and a weight decay of 0.01. For the BERT-based models, the batch size is 20, the learning rate is $3\times 10^{-5}$, and the training is performed for 25 epochs. For NLI-SAN, the size of the hidden layer is 50, the batch size is 30, the learning rate is $10^{-4}$, and the training is performed for 200 epochs.

\subsection{Results and Discussion}
Table \ref{best_results} presents the results of our experiments in terms of macro-averaged F1-score.  Macro F1 score is a suitable metric to evaluate performance on this dataset due to class and fold size imbalance.

In these experiments it is not our goal to outperform state-of-the-art results on the PHEME dataset, but to demonstrate the effectiveness of incorporating the evidence for social media rumour verification.  State-of-the-art results are obtained by more complex architectures, in which incorporating the evidence and evaluating its effects is a more challenging task. For instance, the VRoC model \citep{vroc} currently yields state-of-the-art F1 score of 0.484 on this task, it uses Variational Autoencoder for representation of the rumour as well as multitask learning set up incorporating four tasks.

The results in Table \ref{best_results} suggest that there is indeed a benefit to using the evidence which we have retrieved for rumour veracity classification. This joint approach outperforms the other two, and the use of the rumour alone generally outperforms the use of evidence alone, fitting with the idea that veracity can be classified to some extent by the writing style of the rumour alone. 

In addition to the improvement in the results obtained by having evidence relevant to each rumour, our work opens the door to the use of more complex veracity classification models that consider additional attributes between both elements. The results obtained in the case of the NLI-SAN model show how this approach can be useful, obtaining better results than using the BERT model, although in this case inferior to the more simple use of RoBERTa.

A more detailed, per-class and per-fold, results breakdown for all of the models can be found in Table \ref{best_results}. 
For both BERT and RoBERTa, the combination of rumour together with evidence seems particularly useful for correct classification of the \emph{False} class, with a mild gain also noted for \emph{Unverified}. This could be the result of models inferring that there is disagreement between \emph{False} rumours and their evidence, which would not be possible without the presence of both sources. It is noteworthy that existing rumour veracity classification models using the PHEME dataset have often found the False and Unverified classes to be problematic \citep{SAVED}. \emph{True} class also benefits from incorporating evidence in RoBERTa model comparing to using rumour only. The results breakdown for the NLI-SAN model can also be found in Table \ref{best_results}, for which a similar pattern of per-class results can be observed.
Most of the per-fold results for both BERT and RoBERTa also show the best performance when using a combination of rumour and evidence, only with exception of Germanwings Crash event (dominated by \emph{False} class) for BERT and Ottawa shooting event (dominated by \emph{True} class) for RoBERTa. 

\section{Conclusions and Future Work}

After experimentation with various searching strategies for retrieving evidence from the web, we have constructed the PHEMEPlus dataset, which will facilitate further work on using evidence from wide range of sources for rumour veracity classification. The best such strategies, according to our evaluation metrics, are those which leave the grammatical structure of the claim relatively intact. There is much potential to improve existing rumour veracity classification systems by augmenting them with, or with a broader range, or better quality of evidence. We plan to build upon these findings in the future, working on identifying ways of incorporating the evidence from heterogeneous sources into more complex rumour verification models to maximise the gains from this information and achieve state-of-the-art results.


\section{Acknowledgements}
This work was supported by an EPSRC grant (EP/V048597/1). JDL was funded by the EPSRC Doctoral Training Grant. ML and YH are supported by Turing AI Fellowships (EP/V030302/1, EP/V020579/1).
\bibliography{bib}
\bibliographystyle{acl_natbib}




\end{document}